\documentclass[runningheads]{llncs}
\usepackage[T1]{fontenc}
\usepackage{graphicx,verbatim}
\usepackage{booktabs}
\usepackage{bbm}
\usepackage{amsfonts}
\usepackage{amsmath}
\usepackage{marvosym}
\usepackage{multirow}
\usepackage[colorlinks=true, citecolor=blue, linkcolor=blue]{hyperref}

\begin{document}
\def\eg{\emph{e.g.}}
\def\ie{\emph{i.e.}}
\def\cf{\emph{c.f}\onedot} \def\Cf{\emph{C.f}\onedot}
\def\etc{\emph{etc.}}
\def\wrt{w.r.t\onedot} \def\dof{d.o.f\onedot}
\def\etal{{\em et al.}}

\def\ourmodel{A2ONet}

\title{Attenuation-Resilient Alternating Optimization for Laparoscopic Liver Landmark Detection}

\titlerunning{\ourmodel~for Laparoscopic Liver Landmark Detection}
%

\author{
Lanqing Liu\inst{1}, 
Ruize Cui\inst{1}, 
Jialun Pei\inst{2}\textsuperscript{(\Letter)}, 
Diandian Guo\inst{2}, 
Tiffany Y. So\inst{2}, 
Pheng-Ann Heng\inst{2}, 
and Jing Qin\inst{1} 
}
\authorrunning{L. Liu et al.}
\institute{
The Hong Kong Polytechnic University, Hong Kong, China \\
\and
The Chinese University of Hong Kong, Hong Kong, China \\
\email{peijialun@gmail.com}
}
  
\maketitle              
\begin{abstract}
Liver surface landmark detection is a fundamental prerequisite for anatomical guidance in laparoscopic liver surgery. However, it remains unreliable in practice due to two pervasive challenges: illumination attenuation in underexposed regions and the structural mismatch between pixel-wise localization and continuous curvilinear geometry. 
To address these limitations, we propose~\ourmodel, an attenuation-resilient alternating optimization network for robust liver landmark detection.
To mitigate illumination attenuation, \ourmodel~embraces an illumination field compensation (IFC) block that adaptively enhances dark regions while preserving structural consistency. Meanwhile, we introduce a lightweight frequency–orientation selective filter (FOSF) to suppress repetitive texture interference and preserve salient curvilinear cues. 
Building upon these resilient representations, we design an alternating seg–curve optimization (ASCO) decoder that iteratively couples dense segmentation with explicit curve modeling, enabling mutual guidance to optimize both structural continuity and endpoint localization.
Extensive evaluations on L3D-2K, L3D, and P2ILF demonstrate consistent improvements over competitive methods, establishing a more reliable foundation for intraoperative anatomy guidance.
Our code is available at \href{https://github.com/hyperiondk115/A2ONet}{https://github.com/hyperiondk115/A2ONet}.


\end{abstract}

\section{Introduction}
Laparoscopic liver surgery offers substantial benefits in reduced trauma and faster postoperative recovery~\cite{nguyen2011comparative,guo2025surgical}.
However, the restricted field of view and absence of tactile feedback significantly complicate intraoperative anatomical orientation.
To mitigate these limitations, augmented reality (AR) guidance has been increasingly explored to link intraoperative laparoscopic views with preoperative CT/MRI-derived liver models~\cite{koo2022automatic,ali2025objective,zhou2025landmark}.
Reliable detection of anatomical landmarks on the liver surface, \ie, ridges, falciform ligaments, and silhouette boundaries, serves as semantic anchors for establishing 2D–3D correspondence and supporting accurate registration~\cite{plantefeve2014automatic}.
Consequently, anatomically consistent and geometrically reliable landmark detection is fundamental for robust AR-guided laparoscopic interventions.

\begin{figure}[t]
\includegraphics[width=\textwidth]{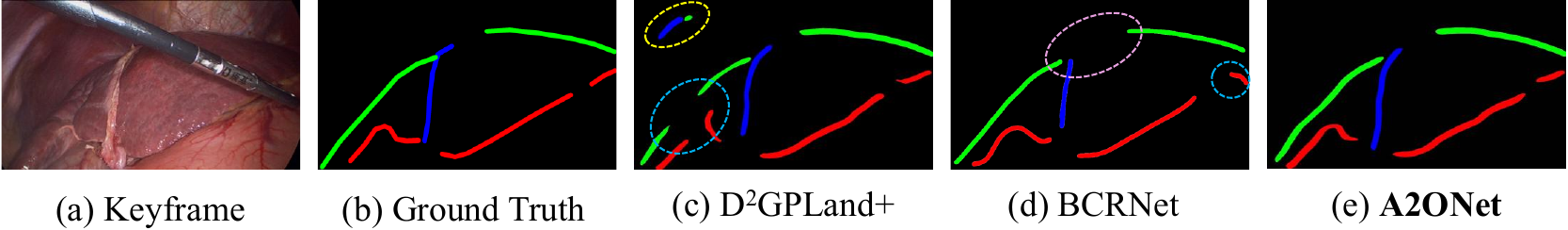}
\caption{Illustration of key challenges for liver landmark detection. Blue circles indicate errors caused by illumination attenuation, yellow circles denote segmentation false positives, and pink circles mark inaccurate endpoint localization.}
\label{fig:introduction}
\end{figure}

Most existing studies formulate liver landmark detection as a dense segmentation problem~\cite{ronneberger2015u,xiao2018weighted,ali2025objective}.
In real surgical scenarios, however, there exists a structural mismatch between pixel-level targets and the continuous-curve geometry of anatomical landmarks.
Although mask-based segmentation methods achieve remarkable performance in pixel-level localization, they often fragment delicate curvilinear landmarks into disconnected pieces, yielding local false positives. 
Recent studies~\cite{pei2024depth,cui2026depth} leverage large vision models~\cite{kirillov2023segment} to enhance detection robustness and combine with the depth modality to augment structural cues~\cite{cui2025topology}.
Despite this progress, these methods remain unable to overcome the impacts of illumination attenuation in peripheral or far-field regions on laparoscopic imaging, which result in blurred landmark boundaries and reduced contrast of surrounding tissue structures (see Fig.~\ref{fig:introduction}).
Beyond light degradation, another challenge lies in balancing positioning accuracy with structural continuity. While incorporating curve geometry modeling enhances landmark continuity~\cite{li2025bcrnet}, it may hinder precise endpoint localization due to excessive smoothing priors. Accordingly, overcoming the ``localization-continuity tradeoff’’ is crucial for stabilizing the consistency of predicted liver landmarks.

In this work, we present~\ourmodel, an Attenuation-resilient Alternating Optimization Network that unifies appearance compensation and structure-aware refinement within a unified framework. 
To address illumination attenuation, we introduce an \textit{illumination field compensation} (IFC) block that adaptively restores underexposed regions while preserving structural coherence. 
Moreover, a lightweight \textit{frequency–orientation selective filter} (FOSF) is embedded to suppress texture-driven ambiguities and enhance salient curvilinear structures. 
Building upon these attenuation-resilient representations, we further design an \textit{alternating seg–curve optimization} (ASCO) decoder that couples a segmentation branch for dense localization and a curve branch for continuous geometry. 
ASCO iteratively injects curve priors to regularize fragmented segmentation responses and uses refined segmentation features to guide subsequent curve updates, resulting in cascaded and mutually reinforcing optimization.
The entire model is trained end-to-end with joint supervision over masks, curves, and illumination conditions.
Extensive experiments on three benchmarks, L3D~\cite{pei2024depth}, L3D-2K~\cite{cui2026depth}, and P2ILF~\cite{ali2025objective} demonstrate consistent improvements over state-of-the-art methods, \eg, reaching 53.51\% DSC and 38.32\% IoU on the L3D-2K dataset, establishing a more reliable foundation for intraoperative anatomical guidance. 

\begin{figure}[t]
\centering
\includegraphics[width=0.95\textwidth]{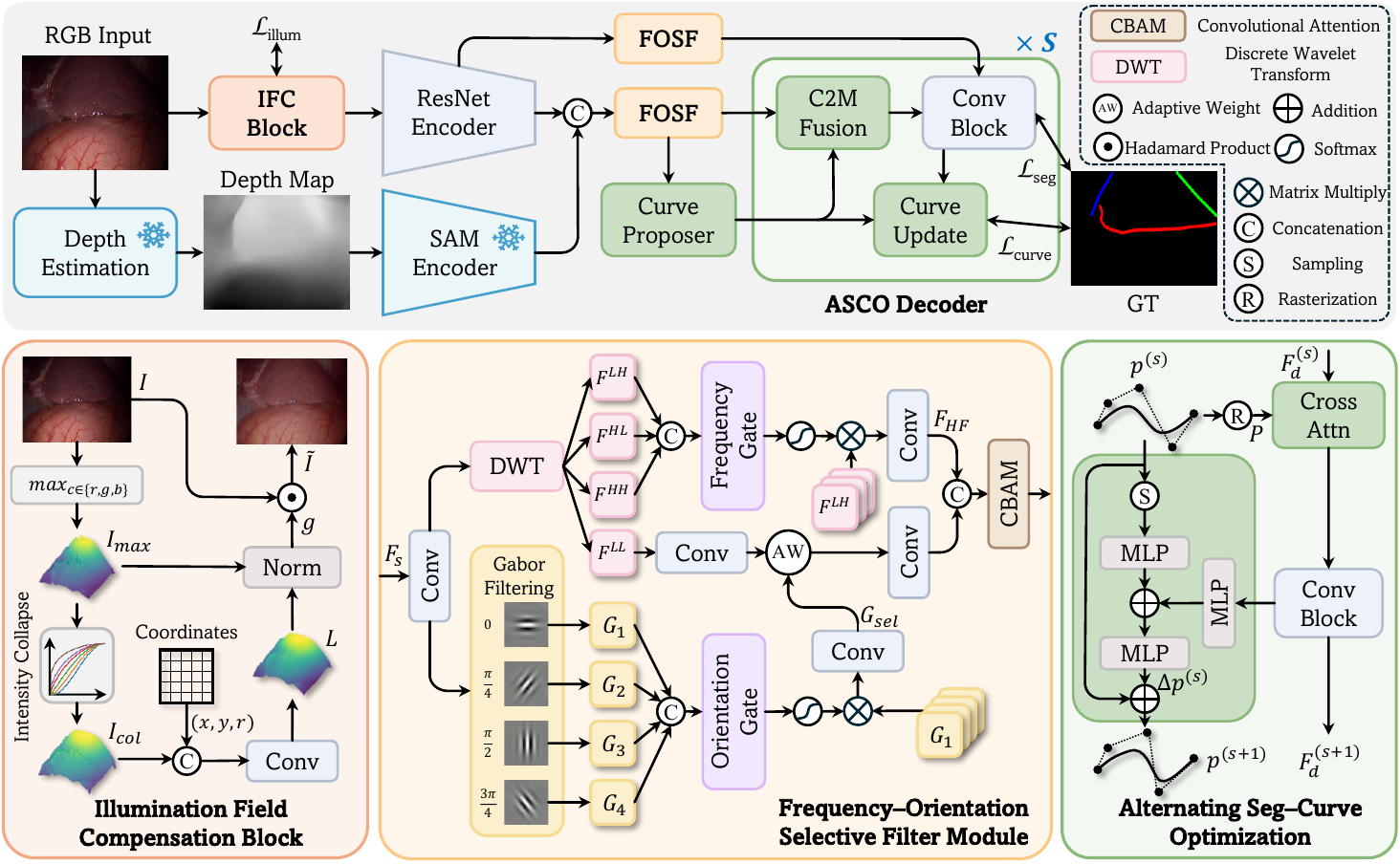}
\caption{Overall architecture of proposed~\ourmodel~for liver landmark detection.}
\label{fig:overview}
\end{figure}

\section{Methodology}
As illustrated in Fig.~\ref{fig:overview}, the proposed \ourmodel~is composed of four key components: an illumination field compensation (IFC) block, a multi-modal encoder, a lightweight frequency–orientation selective filter (FOSF), and a dual-branch alternating seg–curve optimization (ASCO) decoder.
Given an input RGB keyframe,~\ourmodel~first adopts the IFC block to normalize varying illumination for mitigating peripheral attenuation and then employs a ResNet~\cite{he2016deep} to capture appearance representations from the compensated RGB image.
Concurrently, following the design of D2GPLand+~\cite{cui2026depth}, Depth Anything V2~\cite{yang2024depth} estimates a monocular depth map from the RGB input to supply global geometric cues complementary to visual appearance~\cite{pei2024depth}, followed by a frozen SAM image encoder~\cite{kirillov2023segment} to extract depth-aware geometry features. 
Subsequently, we concatenate RGB-D features and utilize FOSF modules to operate on the bottleneck features, emphasizing curvilinear structures while suppressing repetitive textures. 
We further include a curve proposer to initialize coarse parametric curves from the deepest FOSF-enhanced feature.
Building on these initial curves, the ASCO decoder integrates a segmentation branch with FOSF-based skip connections for dense localization and a curve branch for explicit landmark geometry parameterization.
In successive refinement stages, injected curve priors regularize segmentation responses through curve-to-mask (C2M) fusion, and the mask features in turn guide curve updates.

\subsection{Illumination Field Compensation Block}
To enhance detection robustness under low-light laparoscopic environment, we design an IFC block that explicitly normalizes spatially varying illumination.
Given an input RGB image $I\in\mathbb{R}^{H\times W\times 3}$, we adopt a standard reflectance-illumination decomposition assumption: $I(x)=R(x)\odot{L}(x)$, where $L(x)$ denotes a spatial illumination field and $R(x)$ represents illumination-invariant reflectance.
Our objective is to estimate a smooth low-frequency $L(x)$ and compensate illumination while preserving structural consistency.
We first derive a collapsed intensity map from channel-wise maximum: $I_{max}(x)=\max_{c\in\{r,g,b\}}I_c(x)$, followed by an adaptive intensity collapse:
\begin{equation}
I_{\mathrm{col}}(x) = \left(\sin\!\left(\frac{\pi I_{max}(x)}{2}\right)+\epsilon\right)^{\tfrac{1}{k}}, k>0,
\end{equation}
where $k$ is a learnable temperature parameter that modulates the dynamic-range compression intensity, enabling stable illumination estimation while maintaining details.
The transformed intensity is concatenated with normalized spatial coordinates $(x,y)$ and radial distance $r$ to estimate $L(x)$, which is then used to normalize intensity: $I_{\mathrm{norm}}(x)=I_{max}(x)/(L(x)+\epsilon)$.
To prevent over-amplification, we derive a residual gain map: $g_{\mathrm{raw}}(x)=I_{\mathrm{norm}}(x)/(I_{max}(x)+\epsilon)$, and apply residual blending in the gain domain: $g(x)=1+\alpha\big(g_{\mathrm{raw}}(x)-1\big)$, where $\alpha\in[0,1]$ controls the compensation strength.
Modulating the original image produces the final illumination-compensated output: $\tilde{I}(x)=I(x)\odot g(x)$ with $g(x)$ bounded to a valid range, producing a illumination-normalized RGB image that is highly invariant to low-light degradation.

\subsection{Frequency–Orientation Selective Filter}
To enhance the perception of curvilinear features while suppressing texture-induced ambiguities, we design the frequency–orientation selective filter (FOSF) module to operate on bottleneck features.
Given an intermediate  feature $F_{\mathrm{s}}\in\mathbb{R}^{H\times W\times C}$, we first construct a single-channel guidance map $x_{\mathrm{in}}\in\mathbb{R}^{H\times W}$ for frequency-orientation analysis.
Then, the Haar wavelet transform~\cite{mallat2002theory} is applied to obtain four frequency sub-bands $\{F^{\mathrm{LL}},F^{\mathrm{LH}},F^{\mathrm{HL}},F^{\mathrm{HH}}\}$.
To retain informative high-frequency components that are highly related to curvilinear features, we perform per-pixel competition among the three high-frequency sub-bands.
Concretely, a lightweight gate is adopted to process $\{|F^{\mathrm{LH}}|,|F^{\mathrm{HL}}|,|F^{\mathrm{HH}}|\}$ and predict normalized weights $(w_{\mathrm{LH}},w_{\mathrm{HL}},w_{\mathrm{HH}})$ via softmax, yielding the aggregated high-frequency response:
\begin{equation}
F_\mathrm{HF} = w_{\mathrm{LH}}|F^{\mathrm{LH}}| + w_{\mathrm{HL}}|F^{\mathrm{HL}}| + w_{\mathrm{HH}}|F^{\mathrm{HH}}|,
\end{equation}
which prevents single texture-heavy band from domination.
To further emphasize elongated structures possessing coherent directions, we apply $T$ fixed-orientation Gabor filters~\cite{mehrotra1992gabor} on $x_{\mathrm{in}}$, producing orientation-specific responses $\{G_j\}_{j=1}^{T}$.
Then, a lightweight gating network predicts pixel-wise orientation weights $\{\pi_j(x)\}_{j=1}^{T}$ from concatenated responses.
The selected directional evidence is computed as
\begin{equation}
G_{\mathrm{sel}}(x) = \sum_{j=1}^{T} \pi_j(x)\, G_j(x),
\end{equation}
thereby adaptively selecting the dominant orientation at each spatial location.
We then balance low-frequency context and oriented structure by predicting mixing weights $(\alpha,\beta)$ conditioned on $\{F^{\mathrm{LL}}, G_{\mathrm{sel}}\}$ to form an adaptive integration that jointly preserves coarse geometry continuity and directional coherence.
The selected high-frequency response $F_\mathrm{HF}$ is concatenated subsequently and the fused features are finally refined by a lightweight attention block~\cite{woo2018cbam} and forwarded to the decoder for subsequent optimization.

\subsection{Alternating Seg–Curve Optimization}
To jointly model dense appearance cues and explicit curvilinear geometry, the alternating seg–curve optimization (ASCO) decoder utilizes two coupled branches to iteratively refine segmentation prediction and curve estimation.
To be specific, we model each landmark as a Bezier curve $C$ parameterized by $M$ control points $\{p_m\}_{m=1}^{M}$, which ensures geometric continuity while enabling differentiable control-point updates.
At the coarsest decoding stage, initial curve hypotheses are generated in an anchor-based manner, where each spatial location provides a normalized coordinates $a(x)\in(0,1)^2$.
For each landmark class $m$, the proposer predicts confidence scores and regresses control-point offsets $\Delta_m(x)$ in logit space: $p_m(x)=\sigma(\mathrm{logit}(a(x))+\Delta_m(x))$, where $\mathrm{logit}(x)=\log\left(x/(1-x)\right)$ with $x$ clamped to $[10^{-6},\, 1-10^{-6}]$ for numerical stability, and $\sigma(\cdot)$ is the sigmoid function.
The top-$K$ sets of anchors are selected as initial geometric hypotheses, providing coarse structural initialization.
To inject curve priors into dense prediction, the predicted curve is softly rasterized into a spatial prior map:
\begin{equation}
P(x) = \exp\!\left(-\frac{1}{2\sigma^2}\min_{t\in[0,1]}\|x-C(t)\|_2^2\right).
\end{equation}
Through cross-attention, the rasterized prior constrains decoder features to align with the underlying curve geometry.
Starting from the coarse proposals, the curve branch undergoes refinement using segmentation features across decoding stages.
At stage $s$, decoder features $F_d^{(s)}$ are sampled along the current curve hypothesis $C^{(s)}$, yielding a curve embedding that predicts control-point updates:
\begin{equation}
p_k^{(s+1)} = p_k^{(s)} + \Delta p_k^{(s)}, \qquad 
\Delta p_k^{(s)} = \mathcal{R}^{(s)}\!\left(\Phi\!\left(F_d^{(s)}, C^{(s)}\right)\right),
\end{equation}
where $\Phi(\cdot)$ denotes feature sampling and $\mathcal{R}^{(s)}(\cdot)$ is a stage-specific refinement head.
Re-rasterizing the refined curve and re-injecting it into the segmentation branch forms a cascaded alternating loop executed $S$ times.

\subsection{Training Objectives}
The network is trained end-to-end with a weighted multi-task objective:
\begin{equation}
\mathcal{L} = \mathcal{L}_{\mathrm{seg}} + \lambda_{\mathrm{curve}} \mathcal{L}_{\mathrm{curve}} + \lambda_{\mathrm{illum}} \mathcal{L}_{\mathrm{illum}} .
\end{equation}
The term $\mathcal{L}_{\mathrm{seg}}$ supervises dense landmark masks using a combination of Dice and BCE losses to balance region overlap and pixel-wise classification.
We further apply a modified center-line loss~\cite{cui2025topology} on the curves to enforce structural continuity of elongated landmarks across refinement stages.
Finally, to stabilize illumination compensation and avoid over-amplification, we regularize both the predicted illumination field and the gain map by penalizing spatial variations and excessive deviation from unity:
\begin{equation}
\mathcal{L}_{\mathrm{illum}} = \|\nabla L\|_1 + \| g - \mathbf{1} \|_1 .
\end{equation}

\section{Experiments}
\subsection{Datasets and Evaluation Metrics}
We evaluate the proposed method on three public laparoscopic liver landmark benchmarks: L3D~\cite{pei2024depth}, L3D-2K~\cite{cui2026depth}, and P2ILF~\cite{ali2025objective}.
Specifically, L3D comprises 1,152 annotated keyframes, partitioned into 921 for training, 122 for validation, and 109 for testing.
L3D-2K includes 2,000 keyframes, split into 1,532/230, and 238 images for training, validation, and testing, respectively.
For P2ILF, which originally provides 183 annotated images with 167 allocated for training, the unavailability of the official testing set necessitates subdividing its training subset into 124 images for training and 43 for testing.
To quantify region-level segmentation accuracy, we utilize the dice score coefficient (DSC) and intersection-over-union (IoU), alongside the average symmetric surface distance (ASSD) to evaluate the boundary localization of elongated landmarks.

\subsection{Implementation Details}
The RGB encoder employs a ResNet-34 backbone, while a frozen SAM-b image encoder extracts depth features.
Geometrically, we use $T=4$ Gabor filters with orientations $\{0, \frac{\pi}{4}, \frac{\pi}{2}, \frac{3\pi}{4}\}$, and each landmark curve is parameterized by $M=5$ control points refined across $S=4$ cascaded stages with $K=16$ initial anchors.
For data preprocessing, input images are resized to $1024\times1024$ and augmented via random rotation and flipping.
The framework is trained on a single NVIDIA GeForce RTX 3090 GPU with a batch size of 2 for 100 epochs.
Optimization proceeds via AdamW with an initial learning rate of $1e-4$ and a weight decay of $1e-5$, guided by a cosine learning rate schedule to reach a minimum learning rate of $1e-6$.
The weighting coefficients are set to $\lambda_{\mathrm{curve}}=0.2$ and $\lambda_{\mathrm{illum}}=0.1$.

\begin{table}[t]
\centering
\scriptsize
\caption{Quantitative comparison on L3D-2K, L3D and P2ILF datasets.}
\label{tab:comparison}
\setlength{\tabcolsep}{4pt}
\begin{tabular}{l|ccc|ccc|ccc}
\toprule
\multirow{2}{*}{Methods} & \multicolumn{3}{c|}{L3D-2K} & \multicolumn{3}{c|}{L3D} & \multicolumn{3}{c}{P2ILF} \\
\cmidrule(lr){2-4}\cmidrule(lr){5-7}\cmidrule(lr){8-10}
 & DSC$\uparrow$ & IoU$\uparrow$ & ASSD$\downarrow$
 & DSC$\uparrow$ & IoU$\uparrow$ & ASSD$\downarrow$
 & DSC$\uparrow$ & IoU$\uparrow$ & ASSD$\downarrow$ \\
\midrule
U-Net~\cite{ronneberger2015u} & 35.17 & 23.74 & 76.78 & 51.39 & 36.35 & 84.94 & 29.89 & 17.89 & 47.95 \\
COSNet~\cite{labrunie2023automatic} & 37.35 & 25.96 & 59.24 & 56.24 & 40.98 & 69.22 & 26.78 & 16.05 & 47.33 \\
Res-UNet~\cite{xiao2018weighted} & 40.39 & 28.43 & 55.80 & 55.47 & 40.68 & 70.66 & 25.84 & 15.76 & 50.97 \\
UNet++~\cite{zhou2019unet++} & 38.70 & 27.26 & 58.70 & 57.09 & 41.92 & 74.31 & 34.96 & 21.57 & 42.39 \\
HRNet~\cite{wang2020deep} & 41.19 & 29.36 & 56.40 & 58.36 & 43.50 & 70.02 & 33.64 & 21.31 & 44.16 \\
DeepLabV3+~\cite{chen2018encoder} & 41.55 & 29.89 & 55.11 & 59.74 & 44.92 & 60.86 & 27.62 & 16.42 & 50.77 \\
TransUNet~\cite{chen2021transunet} & 37.54 & 25.85 & 46.75 & 56.81 & 41.44 & 76.16 & 25.49 & 15.22 & 52.14 \\
CaseNet~\cite{yu2017casenet} & 40.23 & 28.37 & 56.55 & 60.94 & 46.09 & 63.89 & 33.94 & 21.33 & 45.02 \\
nnU-Net~\cite{isensee2024nnu} & 41.25 & 29.30 & 51.28 & 60.97 & 46.14 & 62.03 & 35.41 & 22.06 & 42.64 \\
\midrule
SAM-Adapter~\cite{chen2023sam} & 39.20 & 26.81 & 59.61 & 57.57 & 42.88 & 74.31 & 21.12 & 12.00 & 57.13 \\
SAMed~\cite{zhang2023customized} & 43.72 & 31.20 & 43.01 & 62.03 & 47.17 & 61.55 & 31.73 & 19.42 & 40.08 \\
SAM-LST~\cite{chai2024ladder} & 39.43 & 27.39 & 56.69 & 60.51 & 45.03 & 68.87 & 28.75 & 18.38 & 46.53 \\
AutoSAM~\cite{hu2023efficiently} & 40.47 & 28.57 & 54.58 & 59.12 & 44.21 & 62.49 & 25.71 & 15.43 & 53.65 \\
\midrule
Double Res-UNet~\cite{ali2025objective} & 40.53 & 28.60 & 53.75 & 58.39 & 43.52 & 63.78 & 36.54 & 22.80 & 39.26 \\
D$^2$GPLand~\cite{pei2024depth} & 49.31 & 35.76 & 41.34 & 63.52 & 48.68 & 59.38 & 40.55 & 25.87 & 38.73 \\
D$^2$GPLand+~\cite{cui2026depth} & \underline{51.66} & \underline{37.59} & \underline{36.24} & \underline{64.62} & \underline{50.14} & \underline{57.68} & \underline{42.15} & \underline{27.40} & \underline{37.12} \\
\textbf{\ourmodel} & \textbf{53.51} & \textbf{38.32} & \textbf{30.97} & \textbf{65.31} & \textbf{50.87} & \textbf{49.89} & \textbf{43.02} & \textbf{27.95} & \textbf{33.47} \\
\bottomrule
\end{tabular}
\end{table}

\begin{figure}[t]
\includegraphics[width=\textwidth]{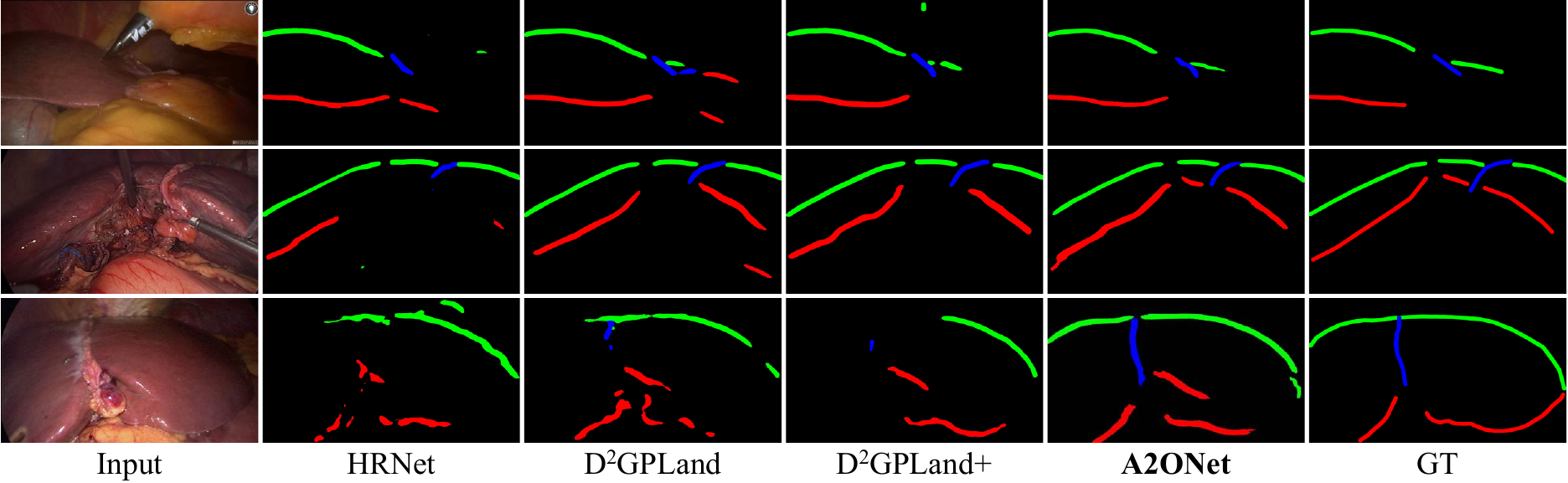}
\caption{Visualizations of the landmark detection results on the L3D-2K, L3D, and P2ILF datasets, arranged from top to bottom respectively.}
\label{fig:visualization}
\end{figure}

\subsection{Comparison with State-of-the-Art}
We benchmark our method against representative baselines classified into three categories: (i) general-purpose segmentation models (U-Net~\cite{ronneberger2015u}, COSNet~\cite{labrunie2023automatic}, Res-UNet~\cite{xiao2018weighted}, U-Net++~\cite{zhou2019unet++}, HRNet~\cite{wang2020deep}, DeepLabV3+~\cite{chen2018encoder}, TransUNet~\cite{chen2021transunet}, CaseNet~\cite{yu2017casenet}, nnU-Net~\cite{isensee2024nnu}), (ii) foundation model-based adaptation methods (SAM-Adapter~\cite{chen2023sam}, SAMed~\cite{zhang2023customized}, SAM-LST~\cite{chai2024ladder}, AutoSAM~\cite{hu2023efficiently}), and (iii) task-specific detection approaches (D$^2$GPLand~\cite{pei2024depth}, D$^2$GPLand+~\cite{cui2026depth}, Double Res-UNet~\cite{ali2025objective}). 
Direct comparison excludes curve-only methods, as their evaluation protocol, \ie, expanding curves to a fixed thickness for region-based assessment, precludes fair benchmarking against mask-based predictors, leaving curve-native comparisons for future work.
Table~\ref{tab:comparison} indicates that our method achieves the highest Dice and IoU across all datasets, establishing robust region-level delineation despite low-light and texture interference.
Compared with D$^2$GPLand+, the strongest competing task-specific baseline, our framework improves DSC/IoU by 1.85\%/0.73\% on L3D-2K, 0.69\%/0.73\% on L3D, and 0.87\%/0.55\% on P2ILF. Notably, it also reduces ASSD by 5.27, 7.79, and 3.65 pixels on the three datasets, respectively, indicating more precise boundary alignment alongside enhanced landmark coverage.
Further, paired significance tests confirm statistically significant improvements across datasets.

Qualitatively, Fig.~\ref{fig:visualization} illustrates that existing baselines frequently generate fragmented predictions along thin curvilinear landmarks under low-light conditions.
Conversely, our method yields continuous, geometrically consistent delineations with fewer false positives, demonstrating the advantages of explicit curve regularization and illumination stabilization.

\begin{table}[t]
\centering
\caption{Ablation study of key components on the L3D-2K dataset.}
\label{tab:ablation}
\setlength{\tabcolsep}{6pt}
\renewcommand{\arraystretch}{0.9}
\begin{tabular}{l|ccccc|ccc}
\toprule
Variant & IFC & $\mathcal{L}_{illum}$ & FOSF & ASCO & $\mathcal{L}_{curve}$ & DSC$\uparrow$ & IoU$\uparrow$ & ASSD$\downarrow$ \\
\midrule
Baseline &  &  &  &  &  & 37.69 & 27.13 & 58.76 \\
V1       & \checkmark &  &  &  &  & 38.23 & 27.76 & 56.90 \\
V2       & \checkmark & \checkmark &  &  &  & 40.45 & 29.04 & 52.36 \\
V3       &  &  & \checkmark &  &  & 39.98 & 28.29 & 53.49 \\
V4       & \checkmark & \checkmark & \checkmark &  &  & 42.48 & 31.14 & 47.82 \\
V5       & \checkmark & \checkmark & \checkmark & \checkmark &  & 44.45 & 33.10 & 43.33 \\
\textbf{\ourmodel} & \checkmark & \checkmark & \checkmark & \checkmark & \checkmark & \textbf{45.16} & \textbf{33.76} & \textbf{40.53} \\
\bottomrule
\end{tabular}
\end{table}

\subsection{Ablation Study}
To verify the effectiveness of individual components in our framework, we conduct ablation experiments on the L3D-2K validation set.
The baseline model consists of a ResNet encoder and a pretrained SAM image encoder, followed by a standard CNN decoder.
Integrating the IFC block (V1) establishes consistent gains over this baseline, indicating the importance of explicitly normalizing low-light appearance.
Further introducing $\mathcal{L}_{illum}$ (V2) yields a larger gain with 2.22\% in DSC, 1.28\% in IoU, and 4.54 pixels in ASSD, indicating that overexposure suppression stabilizes illumination compensation.
Independent application of FOSF modules (V3) also improves performance, demonstrating effective suppression of texture-induced ambiguities.
Combining IFC and $\mathcal{L}_{illum}$ with FOSF blocks (V4) produces the largest jump compared to previous variants, validating their strong complementarity.
Building upon these feature-level enhancements, incorporating the ASCO decoder (V5) further improves overlap and boundary accuracy by 1.97\% in DSC, 1.96\% in IoU, and 4.49 pixels in ASSD, justifying the effectiveness of alternating optimization.
Finally, applying $\mathcal{L}_{curve}$ yields the best overall configuration, confirming that explicit curve supervision provides extra geometric regularization beyond architectural coupling.

\section{Conclusion}
This study introduces~\ourmodel, an attenuation-resilient alternating optimization framework for reliable laparoscopic liver landmark detection. By jointly addressing illumination degradation and texture-induced ambiguities, \ourmodel~enhances representation robustness through illumination field compensation and frequency-orientation selective filtering. More importantly, the proposed alternating seg–curve optimization decoder explicitly reconciles dense pixel-wise localization with continuous curvilinear geometry, alleviating the inherent localization–continuity trade-off. 
Extensive experiments on L3D-2K, L3D, and P2ILF demonstrate consistent improvements over strong baselines, while comprehensive ablation studies confirm the complementary effectiveness of our core components.
By yielding structurally consistent landmarks under challenging illumination, \ourmodel~provides a stable and robust prerequisite for AR-guided navigation. \\
\\
\noindent\textbf{Acknowledgments.}
This work was supported in part by a Shenzhen-Hong Kong-Macao Science and Technology Plan Project (Category C Project) under Shenzhen Municipal Science and Technology Innovation Commission (No. SGDX20230821092359002), in part by the Research Grants Council of the Hong Kong Special Administrative Region, China, under Project T45-401/22-N, and in part by a project under the Collaborative Research with World-leading Research Groups scheme of The Hong Kong Polytechnic University (No. G-SACF). \\
\\
\noindent\textbf{Disclosure of Interests.}
The authors have no competing interests.

%
%
%
\bibliographystyle{splncs04}
\bibliography{Paper-1734}

@inproceedings{plantefeve2014automatic,
  title={Automatic alignment of pre and intraoperative data using anatomical landmarks for augmented laparoscopic liver surgery},
  author={Plantefeve, Rosalie and Haouchine, Nazim and Radoux, Jean-Pierre and Cotin, St{\'e}phane},
  booktitle={International Symposium on Biomedical Simulation},
  pages={58--66},
  year={2014},
  organization={Springer}
}

@article{koo2022automatic,
  title={Automatic, global registration in laparoscopic liver surgery},
  author={Koo, Bongjin and Robu, Maria R and Allam, Moustafa and Pfeiffer, Micha and Thompson, Stephen and Gurusamy, Kurinchi and Davidson, Brian and Speidel, Stefanie and Hawkes, David and Stoyanov, Danail and others},
  journal={International Journal of Computer Assisted Radiology and Surgery},
  volume={17},
  number={1},
  pages={167--176},
  year={2022},
  publisher={Springer}
}

@article{ali2025objective,
  title={An objective comparison of methods for augmented reality in laparoscopic liver resection by preoperative-to-intraoperative image fusion from the miccai2022 challenge},
  author={Ali, Sharib and Espinel, Yamid and Jin, Yueming and Liu, Peng and G{\"u}ttner, Bianca and Zhang, Xukun and Zhang, Lihua and Dowrick, Tom and Clarkson, Matthew J and Xiao, Shiting and others},
  journal={Medical image analysis},
  volume={99},
  pages={103371},
  year={2025},
  publisher={Elsevier}
}

@inproceedings{ronneberger2015u,
  title={U-net: Convolutional networks for biomedical image segmentation},
  author={Ronneberger, Olaf and Fischer, Philipp and Brox, Thomas},
  booktitle={International Conference on Medical image computing and computer-assisted intervention},
  pages={234--241},
  year={2015},
  organization={Springer}
}

@inproceedings{xiao2018weighted,
  title={Weighted res-unet for high-quality retina vessel segmentation},
  author={Xiao, Xiao and Lian, Shen and Luo, Zhiming and Li, Shaozi},
  booktitle={2018 9th international conference on information technology in medicine and education (ITME)},
  pages={327--331},
  year={2018},
  organization={IEEE}
}

@inproceedings{kirillov2023segment,
  title={Segment anything},
  author={Kirillov, Alexander and Mintun, Eric and Ravi, Nikhila and Mao, Hanzi and Rolland, Chloe and Gustafson, Laura and Xiao, Tete and Whitehead, Spencer and Berg, Alexander C and Lo, Wan-Yen and others},
  booktitle={Proceedings of the IEEE/CVF international conference on computer vision},
  pages={4015--4026},
  year={2023}
}

@inproceedings{pei2024depth,
  title={Depth-driven geometric prompt learning for laparoscopic liver landmark detection},
  author={Pei, Jialun and Cui, Ruize and Li, Yaoqian and Si, Weixin and Qin, Jing and Heng, Pheng-Ann},
  booktitle={International Conference on Medical Image Computing and Computer-Assisted Intervention},
  pages={154--164},
  year={2024},
  organization={Springer}
}

@article{cui2026depth,
  title={Depth-Induced Prompt Learning for Laparoscopic Liver Landmark Detection},
  author={Cui, Ruize and Si, Weixin and Li, Zhixi and Wang, Kai and Pei, Jialun and Heng, Pheng-Ann and Qin, Jing},
  journal={Medical Image Analysis},
  pages={103940},
  year={2026},
  publisher={Elsevier}
}

@inproceedings{cui2025topology,
  title={Topology-constrained learning for efficient laparoscopic liver landmark detection},
  author={Cui, Ruize and Zhang, Jiaan and Pei, Jialun and Wang, Kai and Heng, Pheng-Ann and Qin, Jing},
  booktitle={International Conference on Medical Image Computing and Computer-Assisted Intervention},
  pages={585--594},
  year={2025},
  organization={Springer}
}

@inproceedings{li2025bcrnet,
  title={BCRNet: Enhancing Landmark Detection in Laparoscopic Liver Surgery via Bezier Curve Refinement},
  author={Li, Qian and Liu, Feng and Yang, Shuojue and Shen, Daiyun and Jin, Yueming},
  booktitle={International Conference on Medical Image Computing and Computer-Assisted Intervention},
  pages={77--87},
  year={2025},
  organization={Springer}
}

@article{yang2024depth,
  title={Depth anything v2},
  author={Yang, Lihe and Kang, Bingyi and Huang, Zilong and Zhao, Zhen and Xu, Xiaogang and Feng, Jiashi and Zhao, Hengshuang},
  journal={Advances in Neural Information Processing Systems},
  volume={37},
  pages={21875--21911},
  year={2024}
}

@inproceedings{he2016deep,
  title={Deep residual learning for image recognition},
  author={He, Kaiming and Zhang, Xiangyu and Ren, Shaoqing and Sun, Jian},
  booktitle={Proceedings of the IEEE conference on computer vision and pattern recognition},
  pages={770--778},
  year={2016}
}

@article{mallat2002theory,
  title={A theory for multiresolution signal decomposition: the wavelet representation},
  author={Mallat, Stephane G},
  journal={IEEE transactions on pattern analysis and machine intelligence},
  volume={11},
  number={7},
  pages={674--693},
  year={2002},
  publisher={Ieee}
}

@article{mehrotra1992gabor,
  title={Gabor filter-based edge detection},
  author={Mehrotra, Rajiv and Namuduri, Kameswara Rao and Ranganathan, Nagarajan},
  journal={Pattern recognition},
  volume={25},
  number={12},
  pages={1479--1494},
  year={1992},
  publisher={Elsevier}
}

@inproceedings{labrunie2023automatic,
  title={Automatic 3d/2d deformable registration in minimally invasive liver resection using a mesh recovery network},
  author={Labrunie, Mathieu and Pizarro, Daniel and Tilmant, Christophe and Bartoli, Adrien},
  booktitle={Medical imaging with deep learning},
  year={2023}
}

@article{zhou2019unet++,
  title={Unet++: Redesigning skip connections to exploit multiscale features in image segmentation},
  author={Zhou, Zongwei and Siddiquee, Md Mahfuzur Rahman and Tajbakhsh, Nima and Liang, Jianming},
  journal={IEEE transactions on medical imaging},
  volume={39},
  number={6},
  pages={1856--1867},
  year={2019},
  publisher={ieee}
}

@article{wang2020deep,
  title={Deep high-resolution representation learning for visual recognition},
  author={Wang, Jingdong and Sun, Ke and Cheng, Tianheng and Jiang, Borui and Deng, Chaorui and Zhao, Yang and Liu, Dong and Mu, Yadong and Tan, Mingkui and Wang, Xinggang and others},
  journal={IEEE transactions on pattern analysis and machine intelligence},
  volume={43},
  number={10},
  pages={3349--3364},
  year={2020},
  publisher={IEEE}
}

@inproceedings{chen2018encoder,
  title={Encoder-decoder with atrous separable convolution for semantic image segmentation},
  author={Chen, Liang-Chieh and Zhu, Yukun and Papandreou, George and Schroff, Florian and Adam, Hartwig},
  booktitle={Proceedings of the European conference on computer vision (ECCV)},
  pages={801--818},
  year={2018}
}

@article{chen2021transunet,
  title={Transunet: Transformers make strong encoders for medical image segmentation},
  author={Chen, Jieneng and Lu, Yongyi and Yu, Qihang and Luo, Xiangde and Adeli, Ehsan and Wang, Yan and Lu, Le and Yuille, Alan L and Zhou, Yuyin},
  journal={arXiv preprint arXiv:2102.04306},
  year={2021}
}

@inproceedings{yu2017casenet,
  title={Casenet: Deep category-aware semantic edge detection},
  author={Yu, Zhiding and Feng, Chen and Liu, Ming-Yu and Ramalingam, Srikumar},
  booktitle={Proceedings of the IEEE conference on computer vision and pattern recognition},
  pages={5964--5973},
  year={2017}
}

@inproceedings{isensee2024nnu,
  title={nnu-net revisited: A call for rigorous validation in 3d medical image segmentation},
  author={Isensee, Fabian and Wald, Tassilo and Ulrich, Constantin and Baumgartner, Michael and Roy, Saikat and Maier-Hein, Klaus and Jaeger, Paul F},
  booktitle={International Conference on Medical Image Computing and Computer-Assisted Intervention},
  pages={488--498},
  year={2024},
  organization={Springer}
}

@inproceedings{chen2023sam,
  title={Sam-adapter: Adapting segment anything in underperformed scenes},
  author={Chen, Tianrun and Zhu, Lanyun and Deng, Chaotao and Cao, Runlong and Wang, Yan and Zhang, Shangzhan and Li, Zejian and Sun, Lingyun and Zang, Ying and Mao, Papa},
  booktitle={Proceedings of the IEEE/CVF International Conference on Computer Vision},
  pages={3367--3375},
  year={2023}
}

@article{zhang2023customized,
  title={Customized segment anything model for medical image segmentation},
  author={Zhang, Kaidong and Liu, Dong},
  journal={arXiv preprint arXiv:2304.13785},
  year={2023}
}

@article{chai2024ladder,
  title={Ladder fine-tuning approach for sam integrating complementary network},
  author={Chai, Shurong and Jain, Rahul Kumar and Teng, Shiyu and Liu, Jiaqing and Li, Yinhao and Tateyama, Tomoko and Chen, Yen-wei},
  journal={Procedia Computer Science},
  volume={246},
  pages={4951--4958},
  year={2024},
  publisher={Elsevier}
}

@article{hu2023efficiently,
  title={How to efficiently adapt large segmentation model (sam) to medical images},
  author={Hu, Xinrong and Xu, Xiaowei and Shi, Yiyu},
  journal={arXiv preprint arXiv:2306.13731},
  year={2023}
}

@article{nguyen2011comparative,
  title={Comparative benefits of laparoscopic vs open hepatic resection: a critical appraisal},
  author={Nguyen, Kevin Tri and Marsh, J Wallis and Tsung, Allan and Steel, J Jennifer L and Gamblin, T Clark and Geller, David A},
  journal={Archives of surgery},
  volume={146},
  number={3},
  pages={348--356},
  year={2011},
  publisher={American Medical Association}
}

@inproceedings{woo2018cbam,
  title={Cbam: Convolutional block attention module},
  author={Woo, Sanghyun and Park, Jongchan and Lee, Joon-Young and Kweon, In So},
  booktitle={Proceedings of the European conference on computer vision (ECCV)},
  pages={3--19},
  year={2018}
}

@article{zhou2025landmark,
  title={Landmark-free preoperative-to-intraoperative registration in laparoscopic liver resection},
  author={Zhou, Jun and Gao, Bingchen and Wang, Kai and Pei, Jialun and Heng, Pheng-Ann and Qin, Jing},
  journal={IEEE Transactions on Medical Imaging},
  year={2025},
  publisher={IEEE}
}

@inproceedings{guo2025surgical,
  title={Surgical workflow recognition and blocking effectiveness detection in laparoscopic liver resection with pringle maneuver},
  author={Guo, Diandian and Si, Weixin and Li, Zhixi and Pei, Jialun and Heng, Pheng-Ann},
  booktitle={Proceedings of the AAAI Conference on Artificial Intelligence},
  volume={39},
  number={3},
  pages={3220--3228},
  year={2025}
}

\end{document}